\documentclass[conference]{IEEEtran}
\IEEEoverridecommandlockouts
\usepackage{cite}
\usepackage{amsmath,amssymb,amsfonts}
\usepackage{algorithmic}
\usepackage{graphicx}
\usepackage{url}
\usepackage{textcomp}
\usepackage{xcolor}
\usepackage{fancyhdr}
\def\BibTeX{{\rm B\kern-.05em{\sc i\kern-.025em b}\kern-.08em
    T\kern-.1667em\lower.7ex\hbox{E}\kern-.125emX}}
\newcommand{\firstpagecopyright}{%
  \thispagestyle{fancy}%            apply only to the current page
  \fancyhf{}%                       clear header/footer
  \fancyfoot[C]{\parbox{\textwidth}{\scriptsize%
    \copyright\ 2026 IEEE. Personal use of this material is permitted.
    Permission from IEEE must be obtained for all other uses, in any current or
    future media, including reprinting/republishing this material for advertising
    or promotional purposes, creating new collective works, for resale or
    redistribution to servers or lists, or reuse of any copyrighted component of
    this work in other works.}}%
  \renewcommand{\headrulewidth}{0pt}%
  \renewcommand{\footrulewidth}{0pt}%
  \setlength{\footskip}{24pt}%
}
\begin{document}

\title{CrossLag: Predicting 
%Major Signals for 
Major Dengue Outbreak Signals with Limited Data and Transformers\\
\thanks{This work was partially supported by NSF grants DMS-2151235, DMS-2219904, DMS-2309520.}
}
\author{\IEEEauthorblockN{
%1\textsuperscript{st} 
Ashwin Prabu}
\IEEEauthorblockA{\textit{Department of Mathematics} \\
\textit{University of California, Irvine}\\
Irvine, CA, USA \\
%United States of America \\
aprabu@uci.edu}
\and
\IEEEauthorblockN{%2\textsuperscript{nd} 
Nhat Thanh Tran}
\IEEEauthorblockA{\textit{Department of Mathematics} \\
\textit{University of California, Irvine}\\
Irvine, CA, USA \\
%United States of America \\
nhattt@uci.edu}
\and 
\IEEEauthorblockN{%2\textsuperscript{nd} 
Jack Xin}
\IEEEauthorblockA{\textit{Department of Mathematics} \\
\textit{University of California, Irvine}\\
Irvine, CA, USA \\
%United States of America \\
jack.xin@uci.edu}
\and
\IEEEauthorblockN{%3\textsuperscript{rd} 
Guofa Zhou}
\IEEEauthorblockA{\textit{ 
Department of Population Health \& Disease Prevention} \\
\textit{University of California, Irvine}\\
Irvine, CA, USA \\
%Irvine, United States of America \\
zhoug@uci.edu} }
%\and
%\IEEEauthorblockN{4\textsuperscript{th} Jack Xin}
%\IEEEauthorblockA{\textit{Department of Mathematics} \\
%\textit{University of California, Irvine}\\
%Irvine, United States of America \\
%jack.xin@uci.edu}
% \and
% \and
% \and
% \and
% \and

\maketitle
\firstpagecopyright
\begin{abstract}
Predicting future signals with rapid changes (e.g. large peaks) remains challenging 
%in general 
on limited and noisy time series data. In this paper, we focus on multi-year dengue data in Singapore, where domain knowledge suggests a correlation between dengue case spikes and related climate data. %However, this 
Due to the stochastic evolution of the variables, relationships are hard to model. 
% The relationship is difficult to capture as it is hidden among variables and evolves stochastically over time. 
We present CrossLag, an environmentally informed transformer that incorporates lagging information among signals into the architecture in a low-parameter regime. The model can be viewed as an extension of VAR using transformer architecture. The goal is to leverage domain knowledge to overcome limited training data while utilizing the versatility of a transformer architecture for efficient learning. In the detection task of major outbreaks of dengue, the model outperforms state-of-the-art baselines and shows the merit of a lightweight domain-informed transformer design.
% This document is a model and instructions for \LaTeX.
% This and the IEEEtran.cls file define the components of your paper [title, text, heads, etc.]. *CRITICAL: Do Not Use Symbols, Special Characters, Footnotes, 
% or Math in Paper Title or Abstract.
\end{abstract}


\section{Introduction}
\label{sec:intro}

The forecasting of public health time series is an important scientific problem: the ability to do so can detect dangerous anomalies, thus helping to make a quick and effective response. Dengue is one of the deadliest vector-borne diseases in a great number of tropical regions in the world. The limited availability of resources to help treat and prevent it in a large number of places gives rise to the importance of being able to predict outbreaks, 
%trends, 
especially when there is a spike in the number of cases of infection. 

Many studies \cite{Zhou2004, Morin2013, Chuang2017, Liyanage2022, insects13020163, MOKHTAR2024, lag_mosq} have already discussed the correlation between climate and oceanic data and vector-borne diseases. The climate in a region plays a crucial role in the propagation of mosquitoes and the rate of development of both vector and virus \cite{Chuang2017}. Additionally, variables based on oceanic data, such as Indian Ocean Dipole, impact the local climate and, in turn, epidemic patterns and trends \cite{Chuang2017}. Though the
relationship between the variables is known conceptually, it is complicated and well 
hidden within data, compounded by the highly non-stationary
and stochastic nature of the variables. Smaller outbreaks may be linked to local factors such as favorable local climate conditions \cite{insects13020163}, while regional outbreaks are likely associated with large-scale climatic events \cite{MOKHTAR2024}. 

Today, many state-of-the-art models used in time series forecasting are transformers \cite{TimeXer, Autoformer, VCFormer}. The attention mechanism \cite{Attn} present in these models has allowed them to capture complicated properties of the data and predict future patterns with high accuracy. However, many transformers require large amounts of data to train on. This poses a challenge in using them for the dengue forecasting task, as dengue data tends to be less abundant than other time series data. For example, \cite{FWin} illustrates the potential of such models to capture trends in the data, but also the difficulty for them to capture the outbreak patterns well. Thus, there is a need for a compact and efficient model that can capture the complex structure while training on a small amount of data. 

TimeXer \cite{TimeXer} is a transformer that explicitly models the interactions between the endogenous feature (target feature) and the exogenous features. However, it does not directly incorporate any domain knowledge: it infers dependencies from the data directly. Prior works \cite{Autoformer, CorAttn, VCFormer} have demonstrated the benefit of introducing a lag-based calculation within attention, in a variety of ways. However, the lags tend to be selected within the model by computation. Although logical for the general case, this does not fully utilize known lagged effects in the specific field to develop a lightweight model: for example, the lagged effects of temperature and precipitation in a region on mosquito populations as observed in \cite{lag_mosq}. 

In this paper, we introduce CrossLag, a lightweight transformer that utilizes higher-order embeddings and cross-attention to allow for the interaction of the endogenous feature with {\it each of the exogenous features placed in fixed lags based on domain knowledge of mosquito development.} We study the model’s performance in detecting large spikes in the Singapore dengue data with the aid of selected regional climate and oceanic data. 

The main contributions are:
\begin{itemize}
    \item A per-feature embedding that encodes domain-relevant information such as weekly periodicity, annual drift, and local properties to enrich the embedding.
    \item A variant of attention that matches each query with a subset of recent keys ($L$ out of $T$ keys), reducing computation cost while preserving relevant information.
    \item A model that utilizes the aforementioned embeddings and attention to allow for the interaction of the endogenous feature with each of the exogenous features of fixed lags based on domain knowledge.
    \item We study the model's performance in forecasting outbreaks
    %trends 
    in the dengue data from Singapore with the aid of selected regional climate and oceanic data, demonstrating its effectiveness and potential for the task.
\end{itemize}

\section{Data}
\label{sec:Data}

\begin{figure}[t]
\begin{minipage}[b]{\linewidth}
  \centering
  \centerline{\includegraphics[width=0.6\linewidth]{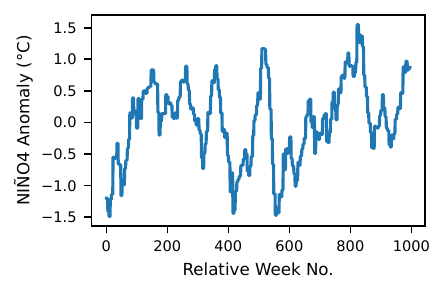}}
%  \vspace{1.5cm}
  \centerline{(a) NI\~NO4 Anomaly}\medskip
\end{minipage}
\begin{minipage}[b]{\linewidth}
  \centering
  \centerline{\includegraphics[width=0.6\linewidth]{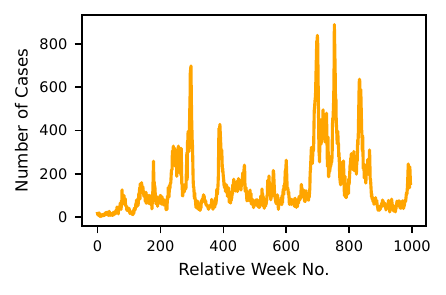}}
%  \vspace{1.5cm}
  \centerline{(b) Cases per Week}\medskip
\end{minipage}
\caption{NI\~NO4 Anomaly and number of cases in weeks.}
\label{fig:Anom_4}
\end{figure}

The dataset comprises weekly dengue data from Singapore, from 2000 to 2019. The data contains close to 1000 samples and 16 features: number of cases, average temperature, precipitation, Southern Oscillation Index (SOI), Oceanic Niño Index (ONI), ONI anomaly, Indian Ocean Dipole (IOD), IOD East, NIÑO1+2, NIÑO3, NIÑO3.4, NIÑO4, and each NIÑO's anomaly. Of these, we use only 5 of the 15 climate and oceanic features: ONI anomaly, NIÑO4 anomaly, IOD East, average temperature, and precipitation. We do this because many oceanic features are highly correlated; thus, we select only a relevant subset of them to represent the influence of oceanic variables on dengue spread. 
% We selected these 3 oceanic features based on their proximity to Singapore and their estimated Mutual Information (MI) \cite{KraskovMI, scikit} with respect to the number of cases. We present a table with the estimated MI of each selected feature with respect to the number of cases in Tab. \ref{tab:MI_table}. 
We also present the plot, Fig. \ref{fig:Anom_4}, of the NIÑO4 anomaly alongside the plot of the number of cases in order to help visualize the data and the reason for feature selection better. \\
% \begin{table}[t]
% % \begin{minipage}[b]{1.0\linewidth}
%     \centering

%     \begin{tabular}{lr}
%     \hline 
%     Feature & Estimated MI\\
%     \hline
%          SOI& 0.28\\
%          ONI & 0.38\\  ONI Anomaly & 0.38 \\IOD & 0.29\\
%          IOD East &0.44\\ NIÑO1+2 & 0.35\\ NIÑO1+2 Anomaly & 0.42\\  NIÑO3 & 0.30\\ NIÑO3 Anomaly& 0.35 \\
%          NIÑO4 & 0.38\\ NIÑO4 Anomaly & 0.48\\  NIÑO3.4 & 0.43 \\ NIÑO3.4 Anomaly & 0.42\\
         
%     \end{tabular}
%         \caption{Estimated MI between each feature and Cases.}
%         \label{tab:MI_table}
% % \end{minipage}
% \end{table}

The split ratio is 6:2:2 (train:val:test) in chronological order.  The data is normalized based on the training set statistics before feeding the input to the models: each feature has a mean of 0 and a standard deviation of 1. Additionally, we scale week stamps uniformly and standardize the year stamps prior to feeding these time stamps into the models. The processed data is available upon request.

% Below is an example of how to insert images. Delete the ``\vspace'' line,
% uncomment the preceding line ``\centerline...'' and replace ``imageX.ps''
% with a suitable PostScript file name.
% -------------------------------------------------------------------------

\section{Model}
\label{sec:pagestyle}

There is variation between the latency of effect for the different features on the number of cases. However, domain knowledge allows us to recognize that certain weeks play more important roles than others. Therefore, standard attention, which looks at all keys and then learns which weeks are important, may fail to capture this, especially when there is a limited amount of data and there is a large amount of noise. Therefore, we would like to incorporate some amount of domain knowledge into our attention, allowing the model to look at a smaller and more relevant subset of time steps for the different features. 

Another important part of the model is the embeddings: they must be used effectively. For example, using the traditional sinusoidal encoding would not be representative of the structure of the data, as it would not reflect the seasonality of the features and the latency of the effects. 

\begin{figure}[t]
\begin{minipage}[b]{1.0\linewidth}
  \centering
  \centerline{\includegraphics[width=5cm, height=6cm]{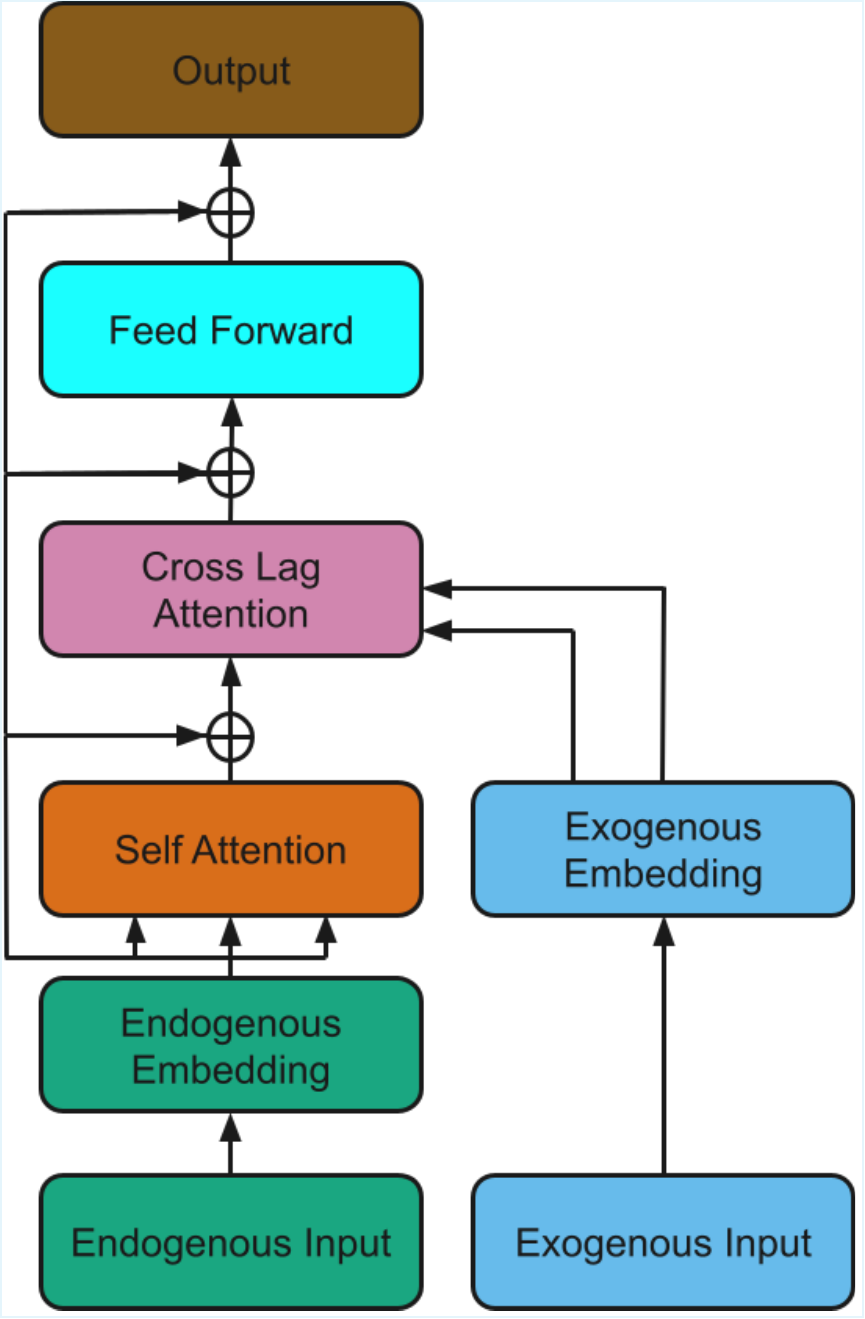}}
\end{minipage}
\caption{Architecture of CrossLag.}
\label{fig:model}
\end{figure}

To ensure that the model makes as much use of domain knowledge and the natural structure of the data as possible, we propose the following architecture for our model: custom embeddings for exogenous and endogenous features, a self-attention on the endogenous feature, a cross attention which incorporates different fixed lags for the keys, and finally a fully connected layer consisting of Multi-Layer Perceptrons (MLPs). Additionally, we regulate the flow of data by using gates that take a weighted sum of the input and output of a layer before passing it onto the next, as seen in Fig. \ref{fig:model}. There are 3 gates present in the model, learnable parameters, to ensure that minimal information loss occurs in the model.

\subsection{Embeddings}
\label{ssec:emb}

To extract the maximum amount of information from the data before passing it on, we encode the following information before projecting to the final embedding: weekly periodicity, annual drift, local average, rate of change, and residual. 

Let $w \in \left[-\frac{1}{2}, \frac{1}{2}\right]^T$ represent the normalized week vector and $y \in \mathbb{R}^T$ the normalized year vector. Let $\theta_i = \frac{2\pi w_i}{P}$ where $P$ represents the periodicity factor, and is $50$ by default. Then, $\forall i\in [1,T]$, $w_{i}^{enc} = \left\{\sin\left(\theta_i\right), \cos\left(\theta_i\right), \sin\left(2\theta_i\right), \cos\left(2\theta_i\right)\right\}$ is the encoded weekly periodicity, and $y_i^{enc} = cy_i+b$ is the encoded annual drift. Note, $c,b$ are learnable scalars. 

Let $x \in \mathbb{R}^T$ be any feature. We compute the rate of change, $x' \in\mathbb{R}^T$ by $x'_t = x_t - x_{t-1} \, \forall t \in [2,T]$ and $x_1' = 0$. 

We compute the local average using a 1D convolution for the previous $N$ values (with padding), set to $4$ by default. We let $\beta \in \mathbb{R}^N$ be the constant convolution weights. Thus, our local average is 
\begin{equation*}
    x^{(m)}_t = \sum_{i=1}^N\beta_ix_{t-i+1} \quad \forall t\in [1,T].
\end{equation*}
Using this information, we also compute the residual $x^{(r)} = x-x^{(m)}$. For the endogenous feature, $x \in \mathbb{R}^T$,
\begin{align}
    X^{enc} &= (x\|y^{enc}\| w^{enc} \| x^{(m)} \| x^{(r)} \| x') \nonumber\\
    X^{endo}&= \text{Dropout}(\text{Linear}(X^{enc})) \in \mathbb{R}^{T \times d_m}
\end{align}
is the endogenous embedding ($\|$ represents concatenation). For each exogenous feature $Z_{:,f}$ where $f \in [1,F]$, we have 
\begin{align}
    Z_{:,f}^{enc} &= (Z_{:,f}\|y^{enc}\| w^{enc} \| Z_{:,f}^{(m)}  \| Z_{:,f}') \nonumber \\
    Z_{:,f, :}^{exo}&= \text{Dropout}(\text{Linear}(Z_{:,f}^{enc})+e_f) \in \mathbb{R}^{T \times d_m}.
\end{align}
where $e_f$ is learnable and initialized to be mutually orthogonal.
We compute the residual for the endogenous feature only. We do this to give the endogenous feature, our target feature, more context, as the model is expected to learn more from it. We perform the exogenous embedding per feature; thus, our result is a 3D tensor $Z^{exo} \in \mathbb{R}^{T \times F \times d_m}$ instead of the usual 2D matrix. We do so to preserve the lag effect of each feature while also being able to enrich the data with the embeddings. This is the key component that allows for the selection of fixed lags in a meaningful manner for each exogenous variable. 

\begin{figure}[t]
\begin{minipage}[b]{1.0\linewidth}
  \centering
  \centerline{\includegraphics[width=1.0\linewidth]{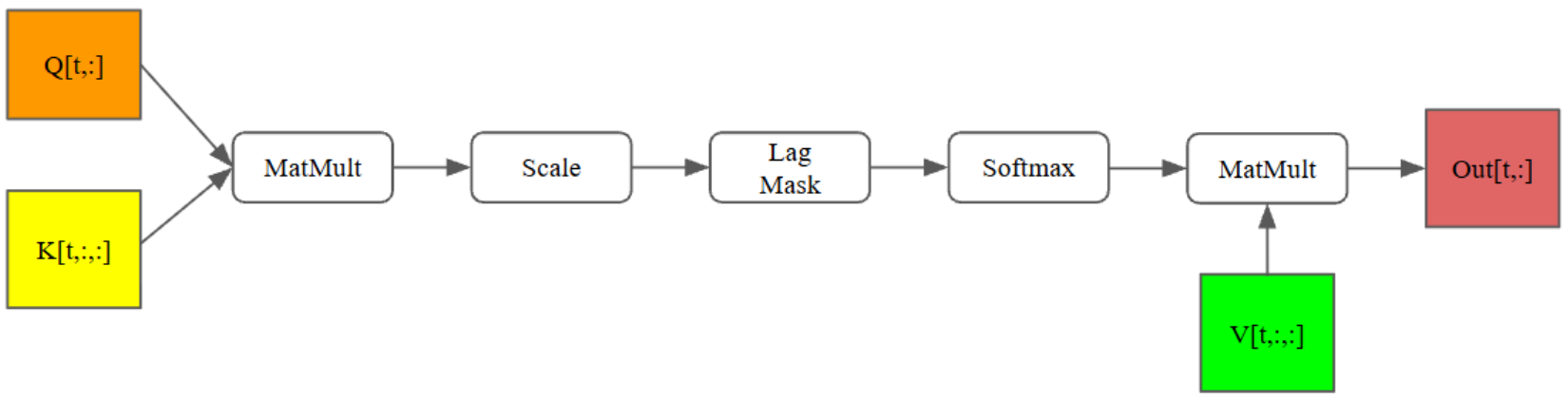}}
\end{minipage}
\caption{CrossLag Attention.}
\label{fig:CLA}
\end{figure}
\subsection{Cross Lag Attention}
\label{ssec:CLA}

The proposed attention mechanism introduces the idea of using a bank of tensors to collect different lagged inputs before projecting them as keys. We use a lag vector $\ell \in \mathbb{Z}_{\geq 0}^L$, with $L$ lag values, $L\ll T$. There are two options offered for the choice of lag vector: an input vector by the user, or a $max\_lag$ hyperparameter, which then selects the lag vector to be $\ell\in \mathbb{Z}_{\geq 0}^L: \ell_i = 2\cdot(i-1), \:\:\forall 1\leq i \leq \lfloor max\_lag/2\rfloor +1$. By default, the model uses $max\_lag = 8$ to generate the lag vector, which is at the responsive end of the lagged effects of climate on dengue spread, keeping it epidemiologically meaningful \cite{Chuang2017}.

Let $L = |\ell|$. For each feature $1\leq f \leq F$ and each lag $1 \leq \delta \leq L$, we create a tensor $Z^{lag} \in \mathbb{R}^{T \times  L \cdot F \times d_{m}}$ where
\begin{equation}
    Z^{lag}_{t, f+(F\cdot (\delta-1)), :} = \begin{cases} Z^{exo}_{t-\ell_\delta, f, :} \; t-\ell_\delta\geq 1,\\ 0 \;\quad\quad \quad  t-\ell_\delta < 1.\end{cases}
\end{equation}

We also define our mask by our valid key set $M(t) = \{f+F\cdot (\delta-1):  t-\ell_\delta\geq 1\} \subset [1,F\cdot L]$.

We have our queries $Q = Linear(X^{endo}) \in \mathbb{R}^{T \times d}$, keys and values $K = V = Linear(Z^{lag}) \in \mathbb{R}^{T \times F\cdot L\times d}$. Thus, our attention mechanism follows the following:
\begin{align}
    S_{t,j} &= \frac{\langle Q_{t,:} , K_{t,j,:}\rangle}{\sqrt{d}}\\
    \hat{S}_{t,j} &= \begin{cases}
        S_{t,j} \quad j \in M(t)\\ -\infty \quad j \not\in M(t)\\    \end{cases} \\
    A_{t,j} &= \begin{cases}
        \frac{\exp\left(\hat{S}_{t,j}\right)}{\sum_{i \in M(t)}\exp\left(\hat{S}_{t,i}\right)} \quad |M(t)| > 0\\
        0 \quad \quad \quad \quad \quad \quad \quad \; |M(t)| = 0
    \end{cases}\\
    O_{t,:} &= \text{Dropout}(A)_{t, : }V_{t, :, :}
\end{align}

The difference between our attention and standard attention is that we fix $t \in [1, T]$ for all $Q,K,V$ instead of allowing the attention to compute for different time indices. This allows the attention to check only certain indices for each query, and reduce 
%. This reduces 
the calculation complexity of the attention scores from  $O(F\cdot T)$ to $O(F\cdot L)$ per query, $L\ll T$.\\

\section{Experiment Setup}
\label{sec:Experiment}
We compare our proposed model, CrossLag, with TimeXer, iTransformer, and VAR on the Singapore data. We use the default parameters of all models in the experiments, adjusting only the encoder input size (6) and model output size (1) in TimeXer and iTransformer. TimeXer and iTransformer use $d_m = 512, d_{ff} = 2048, n_{heads} = 8$ as default, whereas CrossLag has $d_m = 128, d_{ff} = 16, n_{heads} = 1$ as default.
TimeXer and iTransformer use $2$ encoder layers, whereas CrossLag uses only $1$. All transformer models were trained for $30$ epochs, with a patience of $10$ epochs. We use the default optimizer (Adam), initial learning rate (lr=$10^{-4}$), and scheduler for TimeXer and iTransformer; we use AdamW with its default PyTorch parameters and a custom scheduler with an exponential decay (factor = $0.9$) with floor (lr = $5\cdot10^{-5}$) for CrossLag. We use these defaults for each model to ensure that each can perform to the best of its abilities. Finally, we use an input sequence length of 16, a label length of 12, and a prediction length of 24. For VAR, this corresponds to a lag order of 16. All models are trained on CPU. 

% \begin{table}[t]
% % \begin{minipage}[b]{1.0\linewidth}
%     \centering
%     \caption{Hyper-Parameters}
%     \label{tab:MI_table}
%     \begin{tabular}{lcr}
%     \hline 
%     Param & iTransformer & TimeXer & CrossLag\\
%     \hline
%         $d_{\text{model}}$& 512 & 128\\
%         $d_{ff}$& 2048 & 16\\
%         $n\_heads$& 8 & 1\\
         
%     \end{tabular}
% % \end{minipage}
% \end{table}

\section{Experimental Results}
\label{sec:Results}

For this study, we look at the models' ability to predict spikes in dengue cases. Our main goal is to predict the largest spike in the test set. All plots have the number of weeks since the initial input for the x-axis and the
number of dengue cases as the y-axis. The black solid line is the input history available to the models, the blue solid line is the ground truth, and the orange dashed line is the prediction of the models. 

The largest spike in the test set is a jump from approximately 200 cases to 600 cases reported per week, which can be found at index 30 (hence the label ``Truth 30"). Due to the limited availability of data to train on, there are only a few such large spikes for the models to train on, as seen in Fig. \ref{fig:Anom_4}. Therefore, this serves as a test for the models to observe their versatility and how well they can perform when working with limited data.

\begin{figure}[t]
\begin{minipage}[b]{\linewidth}
  \centering
  \centerline{\includegraphics[width=0.6\linewidth]{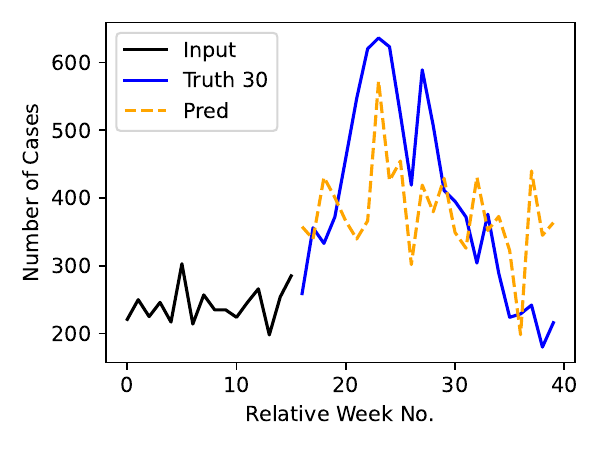}}
%  \vspace{1.5cm}
  \centerline{(a) CrossLag}\medskip
\end{minipage}

\begin{minipage}[b]{\linewidth}
  \centering
  \centerline{\includegraphics[width=0.6\linewidth]{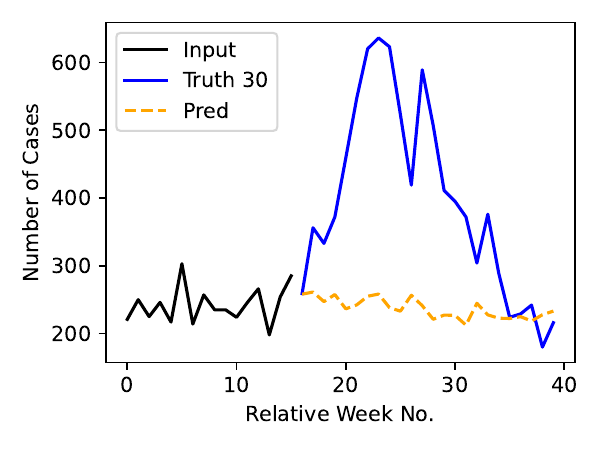}}
%  \vspace{1.5cm}
  \centerline{(b) TimeXer}\medskip
  
\end{minipage}
\begin{minipage}[b]{\linewidth}
  \centering
  \centerline{\includegraphics[width=0.6\linewidth]{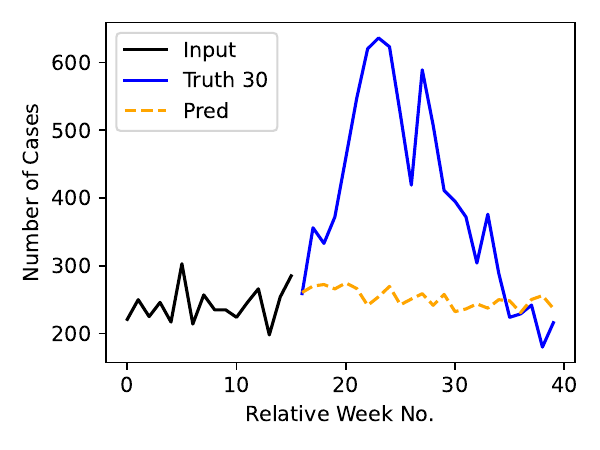}}
%  \vspace{1.5cm}
  \centerline{(c) iTransformer}\medskip
\end{minipage}

\begin{minipage}[b]{\linewidth}
  \centering
  \centerline{\includegraphics[width=0.6\linewidth]{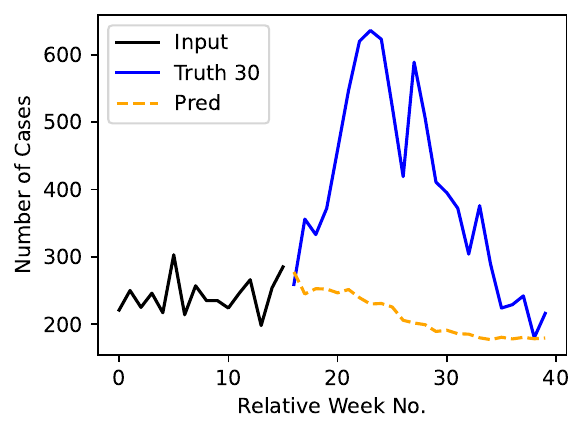}}
%  \vspace{1.5cm}
  \centerline{(d) VAR}\medskip
  
\end{minipage}
\caption{Predictions in default parameter settings.}
\label{fig:default_big}

\end{figure}

% \begin{figure}[t]
% \begin{minipage}[b]{.48\linewidth}
%   \centering
%   \centerline{\includegraphics[width=4.0cm]{default_big_peak_CL.pdf}}
% %  \vspace{1.5cm}
%   \centerline{(a) CrossLag}\medskip
% \end{minipage}
% \hfill
% \begin{minipage}[b]{0.48\linewidth}
%   \centering
%   \centerline{\includegraphics[width=4.0cm]{default_big_peak_TX.pdf}}
% %  \vspace{1.5cm}
%   \centerline{(b) TimeXer}\medskip
  
% \end{minipage}
% \begin{minipage}[b]{.48\linewidth}
%   \centering
%   \centerline{\includegraphics[width=4.0cm]{iTransformer.pdf}}
% %  \vspace{1.5cm}
%   \centerline{(c) iTransformer}\medskip
% \end{minipage}
% \hfill
% \begin{minipage}[b]{0.48\linewidth}
%   \centering
%   \centerline{\includegraphics[width=4.0cm]{VAR.pdf}}
% %  \vspace{1.5cm}
%   \centerline{(d) VAR}\medskip
  
% \end{minipage}
% \caption{Predictions in default parameter settings.}
% \label{fig:default_big}
% %

% \end{figure}

Fig. \ref{fig:default_big} shows a sample of the models' predictions from a random seed. CrossLag can recognize the spike and match the overall pattern with high accuracy. However, iTransformer, TimeXer, and VAR can not recognize that there is a spike in cases. This is explainable as CrossLag is a model that has a greater sensitivity to changes in data by design: the different information encoded in the embeddings provides the model with more information about changes in the data, and the lag selection ensures the model pays attention only to the possibly relevant information. 
%However, this is not a singular result. 

\begin{figure}[tb]

\begin{minipage}[b]{\linewidth}
  \centering
  \centerline{\includegraphics[width=0.6\linewidth]{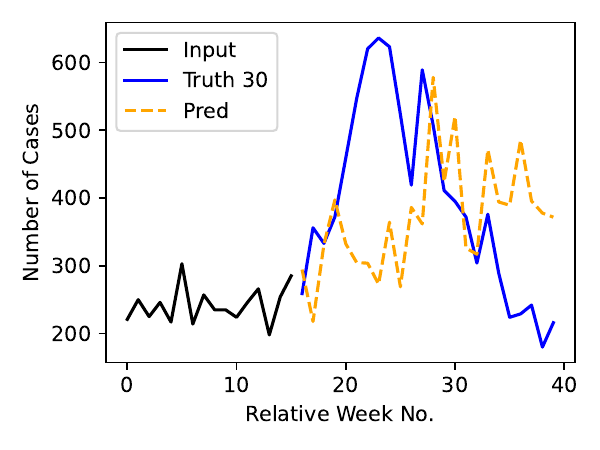}}
%  \vspace{1.5cm}
%  \vspace{1.5cm}
  \centerline{(a) Sample Prediction 1}\medskip
\end{minipage}
%\hfill

\begin{minipage}[b]{\linewidth}
  \centering
  \centerline{\includegraphics[width=0.6\linewidth]{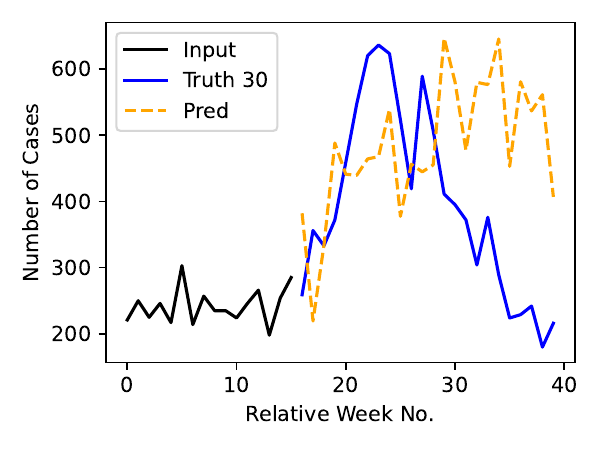}}
%  \vspace{1.5cm}
%  \vspace{1.5cm}
  \centerline{(b) Sample Prediction 2}\medskip
  
\end{minipage}

\caption{Alternate CrossLag predictions from other training seeds.}
\label{fig:crosslag_big}

\end{figure}

Over 10 independent runs, the default CrossLag produces a lower MSE and MAE for predictions just before and during the spike episode (Table~\ref{tab:met}). However, CrossLag has a higher variance, which reflects its sensitive nature. CrossLag is still able to predict that there is a spike in the rare large spike case, which is of great importance as that indicates a more serious period that warrants civilian warning, logistical, medical, and administrative preparations.  
%However,
We note that its predictions 
%are not consistent and 
do not always accurately portray what the spike pattern will be. Fig. \ref{fig:crosslag_big} demonstrates how CrossLag's predictions can vary depending on the random seeds selected in training. CrossLag may not always be able to predict the magnitude of the spike and the timing of the spike exactly.   
%However, this 
This weakness 
does not undermine the fact that CrossLag is effective in providing a trend (not in detail) prediction on small, noisy time series data when certain correlated exogenous factors are identified via domain knowledge. \\

\begin{table}[t]

\caption{Model performances over 10 runs (Raw Case Count).}
\label{tab:met}
% \begin{minipage}[b]{1.0\linewidth}
    \centering
    \begin{tabular}{lll}
    \hline 
    Model (param size) &  MSE (mean $\pm$ std) &   MAE (mean $\pm$ std) \\
    %6326808
    %8451096
    %140364
    \hline
    CrossLag ($\sim 1\times 10^5$)& $\bf{23005 \pm 7052}$ & $\bf{108.71 \pm 17.58}$ \\
        TimeXer ($\sim 8\times 10^6$)& $ 37436 \pm 1873$ & $143.14\pm 4.89 $ \\
        iTransformer ($\sim6 \times 10^{6}$) & $39913 \pm 1348$ & $148.66 \pm 3.98$ \\
        VAR ($\sim 6 \times 10^{2}$) & $37856 \pm 0$ & $138.16 \pm 0$ \\
          \hline
    \end{tabular}

% \end{minipage}
\end{table}

\section{Conclusion}
\label{sec:conc}

We proposed CrossLag, a transformer that takes a domain knowledge-informed approach to data embedding and cross attention. We tested it against two strong baseline transformers, TimeXer and iTransformer, as well as a traditional statistical model, Vector Autoregression (VAR). We compared their performance for forecasting trends in dengue cases in Singapore weeks in advance. CrossLag shows a greater ability to detect dengue outbreaks compared to the other models. Despite its limitations in predicting details, CrossLag offers a new direction for modeling and forecasting vector-borne diseases, utilizing both static domain knowledge and dynamic contextual information present in the data. With further post-processing via statistical methods, CrossLag highlights the potential for compact transformers' use in epidemiology. A future line of work may leverage Fourier transform in CrossLag and enhance its global aspect of climate versus dengue case interaction, as explored in \cite{fwin_2025,AFIDAF}.

% \section{Acknowledgments}
% \label{sec:ack}

\section*{Acknowledgments}
\label{sec:ack}

We would like to thank the Department of Mathematics at UC Irvine for graduate travel support. We 
would also like to 
acknowledge AI assistance (ChatGPT, Claude, and Cursor) in accelerating 
code development. 
Every suggestion and each line of code has been reviewed by the authors to ensure that implementations follow their original thought and intent. The use of AI was primarily for speeding up and refining the process.

\bibliographystyle{IEEEtran}
\bibliography{strings,refs}

\end{document}